\documentclass{article}

\PassOptionsToPackage{numbers, compress}{natbib}
\usepackage[preprint]{ewrl_2024}
\usepackage{nicefrac}
\usepackage{amssymb}

\usepackage[utf8]{inputenc} % allow utf-8 input
\usepackage[T1]{fontenc}    % use 8-bit T1 fonts
\usepackage[hidelinks]{hyperref}       % hyperlinks
\usepackage{url}            % simple URL typesetting
\usepackage{booktabs}       % professional-quality tables
\usepackage{amsfonts}       % blackboard math symbols
\usepackage{nicefrac}       % compact symbols for 1/2, etc.
\usepackage{microtype}      % microtypography
\usepackage{xcolor}         % colors
\usepackage{graphicx}
\usepackage{algorithm}
\usepackage{algpseudocode}
\usepackage{amsmath}

\DeclareMathOperator*{\argmax}{arg\,max}
\newcommand{\method}{SEE}
\newcommand{\methodFull}{Stationary Error-seeking Exploration}

\title{Deterministic Exploration via\\Stationary Bellman Error Maximization}

\author{Sebastian Griesbach$^1$\thanks{Correspondence to \url{sebastian.griesbach@uni-wuerzburg.de}.}~~,
    Carlo D'Eramo$^{1,2,3}$\\
    $^1$ Center for Artificial Intelligence and Data Science, University of Würzburg\\
    $^2$ Department of Computer Science, Technical University of Darmstadt\\
    $^3$ Hessian Center for Artificial Intelligence (Hessian.ai), Germany
    }

\begin{document}

\maketitle

\begin{abstract}
    Exploration is a crucial and distinctive aspect of reinforcement learning~(RL) that remains a fundamental open problem. Several methods have been proposed to tackle this challenge. Commonly used methods inject random noise directly into the actions, indirectly via entropy maximization, or add intrinsic rewards that encourage the agent to steer to novel regions of the state space. Another previously seen idea is to use the Bellman error as a separate optimization objective for exploration. In this paper, we introduce three modifications to stabilize the latter and arrive at a deterministic exploration policy. Our separate exploration agent is informed about the state of the exploitation, thus enabling it to account for previous experiences. Further components are introduced to make the exploration objective agnostic toward the episode length and to mitigate instability introduced by far-off-policy learning. Our experimental results show that our approach can outperform $\varepsilon$-greedy in dense and sparse reward settings.
\end{abstract}

\section{Introduction}
\label{sec:introduction}
Deep reinforcement learning~(RL) algorithms are capable of solving complex tasks such as playing video games only by using pixel inputs~\cite{mnih_playing_2013, van_hasselt_deep_2015}, learning to control intricate bodies in simulation \cite{heess_emergence_2017} and in reality~\cite{openai_learning_2019} or beating humans at complex board games~\cite{silver_mastering_2017, schrittwieser_mastering_2020}. Exploration is a fundamental component of RL, since the ability to solve a given problem comes with the need of discovering a solution in the first place. Many exploration strategies in RL rely on injecting random noise either directly by adding noise to the actions~\cite{sutton_reinforcement_2020,mnih_playing_2013, lillicrap_continuous_2019} or indirectly by maximizing entropy as an additional objective~\cite{schulman_proximal_2017,haarnoja_soft_2018}. Exploration by random noise has been shown to be very effective in many applications. However, it lacks coherence between the taken actions. In an environment with sparse rewards, this can be thought of as performing a random walk in the state space by performing random actions. As random walk is not effective at covering the state space, this method can be ineffective for environments with sparse rewards.

In the setting of sparse rewards, one common approach is to introduce intrinsic rewards to substitute the sparseness of the extrinsic rewards stemming from the environment. Such techniques often rely on state-novelty measures to encourage the agent to explore novel regions of the state space, for example by using generalized visitation counts~\cite{tang_exploration_2017}, discriminating between seen and unseen states~\cite{fu_ex2_2017}, or using prediction errors of random networks~\cite{burda_exploration_2018}. Effectively these methods transform the sparse reward problem back to a dense reward where exploration by random noise is known to work well. With respect to the aforementioned random-walk, this can be thought of as performing a random walk at the edge of the known state space. While these methods have been shown to be effective in sparse reward settings~\cite{burda_exploration_2018}, they have little or even a negative effect in other reward settings~\cite{taiga_bonus-based_2021}. Several conceptual drawbacks to novelty-seeking methods might be the cause of this. For example, indiscriminately seeking state novelty could lead the agent to excessively explore regions of the state space that are not relevant to achieving the goal. Moreover, rewarding novelty results in a non-stationary signal which introduces additional difficulties. By combining extrinsic and intrinsic rewards into a single MDP, the optimal policy might be diluted by the intrinsic reward and negatively affects the performance of the agent.

This work is based on the previously known idea of using the Bellman error as a separate optimization objective for exploration.
Our method \methodFull~(\method) does not rely on random noise and aims to be effective in both dense and sparse reward settings. SEE adopts two separate neural networks, which we call \textit{exploration network} and \textit{exploitation network}, and uses the Bellman error as a proxy measure for interesting transitions. To drive exploration, SEE trains the exploration network by maximizing the absolute Bellman error, approximated with the TD-error of the exploitation network. Crucially, to tackle the problem of non-stationarity of the Bellman error during learning, we use a fingerprinting embedding~\cite{harb_policy_2020} to inform the exploration network about the state of the exploitation network. The exploration network is agnostic to the episode length by maximizing the maximum single TD-error encountered throughout an episode instead of accumulated future TD-errors. To mitigate instability introduced when training too far off-policy~\cite{fujimoto_off-policy_2019} a mixture of both network outputs is used during the rollout. Importantly the resulting exploration policy is deterministic. In the random walk analogy, our method acts such that it selects a random point in a promising state region and tries to move there with coherent actions. In this paper, we conduct a set of experiments sufficing as proof of concept by showing improvements over $\varepsilon$-greedy exploration, as well as an ablation study to verify the effectiveness of all components of the algorithm.

\section{Related works}
\label{sec:related_works}
Most exploration strategies rely on random noise like the $\varepsilon$-greedy strategy employed in DQN \cite{mnih_playing_2013}, or entropy maximization in SAC \cite{haarnoja_soft_2018} or PPO \cite{schulman_proximal_2017}. Some sophisticated exploration methods focus on the case of sparse rewards by introducing an intrinsic reward that encourages exploration by adding a bonus to novel state regions. The underlying mechanic for selection explorative actions stays unchanged. One successful algorithm that follows this concept is Random Network Distillation (RND) \cite{burda_exploration_2018}. RND is designed for large continuous state spaces and therefore can not rely on visitation counts; instead, two additional networks are introduced, a randomly initialized and static target network, and a predictor network aiming to imitate the target network. During training of the agent, encountered states are passed through the target and predictor network. The difference in prediction is used as the intrinsic reward and the predictor network is updated on the data to imitate the target network. This has the effect that the predictor network will be good at imitating the target network at frequently visited states and thus reducing the intrinsic reward, while for rare or new states a high intrinsic reward is expected. The idea of using a predictor error as a reward signal is also a core concept of our method, but instead of using a random target network, the TD-error of the action-value network is used. One known problem of novelty-seeking methods is the so-called "noisy TV problem", which is a thought-experiment in which an agent encounters a screen that will indefinitely show unique images and therefore keep triggering the novelty reward. Our method does not indiscriminately seek novel states but instead aims to maximize the TD-error. Therefore, the noisy TV problem only applies if the transitions also impact the return (see Section \ref{sec:conclusion} for more details).

The idea of using the TD-error as an exploration objective is not a new one \cite{schmidhuber_adaptive_1991, thrun_active_1992}. For example,~\citet{simmons-edler_reward_2020} follow the same conceptual idea as our method of splitting exploration and exploitation into two separate objectives and using the TD-error as the maximization objective for the exploration component. However, the implementation of their method varies greatly from ours, as their method does not include any of the three modifications of our method. To overcome the off-policy problem, which we handle by mixing both policies, they separate the rollouts of the exploitation and exploration agent and save the transitions in separate replay buffers. During the update, the transitions from both buffers are mixeielen Dank.d according to a specific ratio. Their method leverages the idea into the continuous action domain by having two sets of actor and critic networks. For now, we limit our approach to the simpler case of the discrete action spaces to build a solid foundation before moving to continuous control settings.

The exploration-exploitation dilemma is thought of as the conflict of wanting to explore just enough to find the optimal solution and then starting to exploit as soon as possible \cite{sutton_reinforcement_2020}. Unfortunately, in practice we can never be sure about whether the optimal solution has been found. \citet{riedmiller_collect_2021} propose to rethink this paradigm and instead look at exploration and exploitation as separate phases of the process. The exploration phase is considered as an optimization process of gathering an optimal dataset such that an optimal exploitation policy can be learned based on this dataset thereafter. This also means that the data collection process should be aware of what data has already been collected. Only after the exploration phase is done, exploitation starts. Our method roughly follows this concept. We consider the TD-error as a proxy of valuable transitions for the learning process. Through gathering data by maximizing the TD-error, we aim to optimize the usefulness of gathered data. The exploration objective is conditioned on the state of the exploitation objective and thus can take into account what is already known to limit the amount of redundantly gathered data. However, our method also balances exploration and exploitation during the exploration phase.

\section{Preliminaries}
\label{sec:preliminaries}

\subsection{Markov decision processes}
We define a Markov decision process~(MDP)~\citep{puterman_chapter_1990} as a tuple $\langle \mathcal{S},\mathcal{A},\mathcal{P},\mathcal{R},\gamma\rangle$, where $\mathcal{S}$ is a continuous state space, $\mathcal{A}$ is a discrete action space, $\mathcal{P}:\mathcal{S}\times\mathcal{A}\mapsto\mathcal{M}(\mathcal{S})$ is a transition kernel, $\mathcal{R}:\mathcal{S}\times\mathcal{A}\times\mathcal{S}\mapsto\mathbb{R}$ is a deterministic reward function, and $\gamma\in[0,1)$ is a discount factor. A policy $\pi:\mathcal{S}\mapsto\mathcal{A}$ is a probability distribution over actions $a\in\mathcal{A}$ to perform in any state $s\in\mathcal{S}$, which induces an action-value function $Q^\pi(s,a)\triangleq\mathbb{E}\left[\Sigma_{t=0}^\infty\gamma^tR_t|S_0=s,A_0=a\right]$, i.e., the expected cumulative discounted return obtained when executing action $a$ in state $s$ and following policy $\pi$ thereafter. The goal of RL is to find the optimal policy $\pi^*$ corresponding to the one inducing the optimal action-value function $Q^*(s,a)=\max_\pi Q^\pi(s,a)$, which satisfies the Bellman optimality equation~\citep{bellman_theory_1954} $Q^*(s,a)\triangleq\int_\mathcal{S}\mathcal{P}(s'|s,a)[\mathcal{R}(s,a,s')+\gamma \max_{a'}Q^*(s',a')ds'$ and is the fixed point of the optimal Bellman operator $T^*:\mathcal{B}(\mathcal{S}\times\mathcal{A})\mapsto(\mathcal{S}\times\mathcal{A})$ defined as $(T^*Q)(s,a)\triangleq\int_\mathcal{S}\mathcal{P}(s'|s,a)[\mathcal{R}(s,a,s')+\gamma \max_{a'}Q^*(s',a')ds'$.

\subsection{Deep \texorpdfstring{$Q$}{Q}-network}
Deep Q-Network (DQN)~\cite{mnih_playing_2013} is an RL algorithm that leverages a deep neural network to approximate the action-value function $\hat{Q}^\theta(s, a)$, where $\theta$ represents the network parameters. The objective is to minimize the difference between the predicted action values and the target action values derived from the Bellman equation. To stabilize training, DQN employs two key techniques: experience replay and a target network. Experience replay stores the agent's experiences $(s, a, r, s')$ in a replay buffer and samples mini-batches of experiences uniformly at random to update the network, breaking the correlation between consecutive updates. The target network, a copy of the online network, is updated periodically to provide stable target values, reducing the oscillations and divergence during training. The loss function for updating the online network is defined as:

\begin{align}
    \label{eq:exploitation_loss}
    \mathcal{L}_\theta = \mathbb{E}_{s,a,r,s' \sim \mathcal{D}}\left[\left(r + \gamma \max_{a'} \hat{Q}^{\theta'}(s',a') - \hat{Q}^\theta(s,a)\right)^2\right],
\end{align}

where $\langle s, a, r, s'\rangle$ is a transition from the replay buffer $\mathcal{D}$,  $\theta$ and $\theta'$ are parameters of the online and target networks, respectively. In this work, we use a combination of two variants of DQN known as dueling~\cite{wang_dueling_2016} and double DQN~\cite{van_hasselt_deep_2015}. The former aims at improving the estimate of the action-value function by obtaining separate approximations of value function and advantage function, while the latter aims at mitigating the problem of action-value function overestimation typical of DQN by optimizing for the following loss

\begin{align}
    \label{eq:exploitation_loss_double}
    \mathcal{L}_\theta = \mathbb{E}_{s,a,r,s' \sim \mathcal{D}}\left[ \left( r + \gamma \hat{Q}^{\theta'}(s',\argmax_{a'}\hat{Q}^{\theta}(s',a')) - \hat{Q}^\theta(s,a)\right)^2 \right],
\end{align}

\subsection{Fingerprinting}
\label{subsec:fingerprinting}
Fingerprinting has been originally designed as an embedding method for policy networks for the purpose of policy evaluation \cite{harb_policy_2020}. To embed a policy network, a set of probe states is passed through it, and the resulting actions are concatenated and taken as the embedding. Importantly, as this process is fully differentiable, the probe states can be optimized using any gradient-based optimizer. The number of probe states is a hyperparameter, thus the embedding size is independent of the size of the policy network. It has been shown that only a few probe states are needed to carry enough information in an RL setting to learn complex locomotion behavior \cite{faccio_general_2022}. As our method currently is limited to discrete action spaces, we do not have dedicated policy networks, and only action-value networks are used. However, fingerprinting can be applied to embed any neural network without any modification, by concatenating action-value vectors of the probe states instead of actions.

\section{Stationary Error-seeking Exploration}
\label{sec:method}
\label{subsec:exploration_mdp}
We propose a novel exploration strategy for both dense and sparse reward problems, that tackles exploration and exploitation as two separate objectives. Besides the regular RL objective that trains what we call the \textit{exploitation policy}, our method introduces an additional optimization procedure that maximizes the absolute Bellman error to train a separate policy, which we call \textit{exploration policy}. To reduce instabilities common to the optimization of the Bellman error~\cite{sutton_reinforcement_2020}, our approach conditions the exploration network on the exploitation network to render the objective stationary. Because of this, we call our method~\methodFull~(SEE). In the following, we present our method and the required key modifications, namely the use of fingerprinting for conditioning, the use of maximum reward update, and the mixing of objectives to create the deterministic behavior policy.

\subsection{Exploration-exploitation as separate optimization problems}
The exploitation objective is the regular objective of DQN \cite{mnih_playing_2013}, minimizing the loss shown in equation~\ref{eq:exploitation_loss}. Additionally, we use a combination of Dueling~\cite{wang_dueling_2016} and Double DQN~\cite{van_hasselt_deep_2015}. As a second separate component, we want to formulate an exploration objective, leveraging a dedicated approximation network $\hat{\Delta}^\omega$ to predict the absolute TD-error of the exploitation network $\hat{Q}^\theta$.
We propose that our exploration objective solves the MDP of the exploitation objective with two modifications:
\begin{itemize}
\item The parameters of the exploitation network $\theta \in \Theta$ are included in the state space
\begin{align}
    \mathcal{S}_\Delta = \mathcal{S} \cup \Theta;
\end{align}
\item The reward function $\mathcal{R}$ is replaced by the absolute TD-error of the exploitation action-value estimation
\begin{align}
\label{eq:exploration_reward}
    \mathcal{R}_{\Delta}(s,a,s', \theta) = \left| \mathcal{R}(s,a,s') + \gamma \max_{a'} \hat{Q}^{\theta}(s',a') - \hat{Q}^\theta(s,a) \right|,
\end{align}
where $\mathcal{R}(s,a,s')$ is the reward function of the exploitation MDP.
\end{itemize}
Thus, the exploration objective is to maximize the absolute TD-error of the exploitation objective. Note that we are not using a target network here such that this reward entirely depends on the transition and the online parameters $\theta$. In the following, we discuss the aforementioned modifications to the exploration optimization.

\subsubsection{Conditioning of exploration on exploitation}
\label{subsec:conditioning}
During learning, the exploitation action-value approximation changes, causing the TD-errors to be non-stationary. To tackle this, we propose to inform the exploration policy about the cause of these changes. As mentioned above, we include the parameters $\theta$ of the exploitation objective in the state-space of the exploration objective. By doing this, the exploration objective becomes stationary as it is conditioned on the changes in the exploitation value function.
Thus, the exploration value network $\hat{\Delta}^\omega$ receives as input an environment state, an action, and exploitation parameters $\hat{\Delta}^\omega(s, a, \theta)$. Simply taking the parameters as a vector has been proven effective at conveying the necessary information~\cite{faccio_parameter-based_2021}. However, the fingerprinting embedding is scalable as it requires fewer learnable parameters and has an embedding size that can be chosen as a hyperparameter~\cite{harb_policy_2020,faccio_general_2022}. After being passed to the exploration network, the exploitation parameters $\theta$ are embedded by fingerprinting, where the learnable probe states are part of the exploration network parameters $\omega$. The embedding is concatenated with the state vector and passed through the fully connected network to generate the absolute TD-error predictions.

\subsubsection{Maximum reward formulation}
\label{subsec:max_formulation}
Since we consider the absolute TD-error as the exploration reward function $\mathcal{R}_\Delta$, this would lead to a preference for longer episodes, as more positive rewards can be received. Let us consider any environment with a constant negative reward such that the goal is to reach the terminal state as fast as possible. There might be a large TD-error in the transitions towards the terminal state, but through many small TD-errors elsewhere in the state space the exploration agent might be incentivised to prolong the episode. To avoid this undesired behavior, we want to make our exploration policy insensitive to the episode length by seeking the maximum single step TD-error instead of the accumulated one. This can be achieved by the maximum reward formulation of the optimal Bellman update~\cite{gottipati_maximum_2023,veviurko_max_2024}:
\begin{align}
    \label{eq:max_update}
    Q^{k+1}(s,a) = \max\left(r,\gamma\max_{a'}Q^k(s',a')\right).
\end{align}

\subsection{Exploration objective}
The exploration objective is a modified version of DQN including the maximum reward formulation and conditioning as described above. The loss to minimize is
\begin{align}
\label{eq:exploration_loss}
    \mathcal{L}_\omega = \mathbb{E}_{s,a,s' \sim \mathcal{D}, \theta \sim \mathcal{D}_\Theta}\left[\left(\max\left( \mathcal{R}_\Delta(s,a,s',\theta), \gamma\max_{a'}\hat{\Delta}^{\omega'}(s',a', \theta)\right) - \hat{\Delta}^\omega(s,a, \theta)\right)^2\right],
\end{align}
where $\mathcal{D}$ is the transition replay buffer, $\mathcal{D}_\Theta$ is a separate buffer containing the current and previous exploitation objective parameters $\theta$, $\hat{\Delta}^\omega(s,a, \theta)$ is the absolute TD-error approximation network with online parameters $\omega$ and target network parameters $\omega'$. Fingerprinting is used to embed $\theta$. $\mathcal{R}_\Delta(s,a,s',\theta)$ is the exploration reward function defined in Equation~\eqref{eq:exploration_reward}. Note that transitions and exploitation parameters are stored and sampled independently. In practice, we use the double $Q$-learning variant to calculate the target~\cite{van_hasselt_deep_2015}.

\subsection{Mixing of objectives}
\label{subsec:mixing}
In common off-Policy settings, the behavior policy used during rollouts behaves similarly to the learned target policy. For example, it can be a noisy and past variant of the target policy, e.g., when using $\varepsilon$-greedy exploration with a replay buffer. It has been shown that deep off-policy algorithms perform poorly in settings where the behavior policy is too out-of-distribution w.r.t. the target policy~\cite{fujimoto_off-policy_2019}. For this reason, the exploitation policy learns poorly when purely using data generated by the exploration policy. To mitigate this effect we combine both policies to select actions during the rollout, such that the behavior policy effectively is closer in behavior to the exploitation policy.
A weighted average across the values of both objectives selects the action:
\begin{align}
    \label{eq:mixing}
    \mu(s) = \argmax_a \left( (1-\lambda)\hat{Q}^\theta(s,a) + \lambda\hat{\Delta}^\omega(s,a,\theta) \right),
\end{align}
where $\mu$ is the deterministic behavior policy and $\lambda$ is a hyperparameter mixing the exploitation and exploration objectives. In practice, before this deterministic behavior policy comes into play, a warm-up phase takes place to prefill the replay buffer with some transitions. During this warm-up phase, random actions are selected uniformly. Unlike $\varepsilon$-greedy methods, $\lambda$ does not need to be scheduled to decrease over time. As the exploitation objective learns, TD-errors decrease and thus also the impact of the exploration objective. This is not true if there is an unpredictable element in the environment connected to the reward as explained in Section~\ref{sec:conclusion}. Besides the off-policy problem, maximizing the TD-error can cause the exploration to seek unpromising regions. Generally, we do not want to take up much capacity of the networks to accurately describe how bad certain states are if it is clear that they are bad. Balancing exploitation with exploration during this phase helps to steer the gathering of data toward promising state regions.

\begin{algorithm}
\label{alg:see}
\caption{\methodFull~(SEE)}
\begin{algorithmic}
\State \textbf{Input:} exploitation network $\hat{Q}^\theta: \mathcal{S} \times \mathcal{A} \rightarrow \mathbb{R} $ with initialized parameters $\theta$; exploration network $\hat{\Delta}^\omega: \mathcal{S} \times \mathcal{A} \times \Theta \rightarrow \mathbb{R}$
empty transition buffer $D$; empty exploitation parameter buffer $D_\Theta$
\State \textbf{Output:} Learned $\hat{Q}^\theta$
\State
\Repeat
\For{number of environment interactions}
    \State select action according to $a \leftarrow \mu(s)$ \Comment{Equation \eqref{eq:mixing}}
    \State execute $a$ and record transition in transition buffer $D \leftarrow D \cup \{s,a,s',r\}$
\EndFor
\For{number of gradient steps}
    \State sample minibatch from transition buffer $\langle s, a, r, s'\rangle \sim D$
    \State update exploitation parameters $\theta$ by minimizing $\mathcal{L}_\theta$ \Comment{Equation \eqref{eq:exploitation_loss}}
\EndFor
\State add exploitation parameters to parameter buffer $D_\Theta \leftarrow D_\Theta \cup \{\theta\}$
\For{number of gradient steps}
    \State sample minibatch from transition buffer $\langle s, a, r, s'\rangle \sim D$
    \State sample minibatch parameters from parameter buffer $ \theta \sim D_\Theta$
    \State calculate exploration rewards $\mathcal{R}_\Delta$ per transition and parameter combination \Comment{Equation \eqref{eq:exploration_reward}}
    \State update exploration parameters $\omega$ by minimizing $\mathcal{L}_\omega$ \Comment{Equation \eqref{eq:exploration_loss}}
\EndFor
\Until stopping criteria met
\end{algorithmic}
\end{algorithm}

\section{Experimental results}
\subsection{Environments}
\label{subsec:environments}
Many sophisticated exploration algorithms focus on the specific setting of sparse rewards. We are aiming to find a generic exploration mechanism that works in a variety of settings. Therefore, we choose three basic well-known RL problems, implemented in gymnasium~\cite{towers_gymnasium_2023}, that follow different reward schemes. These include a sparse reward environment, a dense reward environment with unshaped reward function, and a dense reward environment with shaped reward function.

For the sparse reward environment, we use a modified version of \texttt{MountainCar-v0}, named \texttt{SparseMountainCar-v0}. In the regular \texttt{MountainCar-v0}, the environment returns a constant reward of $-1$ on each transition to incentivize the agent to reach the terminal state as soon as possible. The agent controls a car with a weak engine and is supposed to climb a hill. To reach the goal it must first gather speed by swinging up and down the valley it is in. To make the environment sparse \texttt{SparseMountainCar-v0} returns a constant reward of $0$ for each transition and a reward of $+1$ upon reaching the terminal state.

For the dense unshaped reward setting, we use unmodified \texttt{CartPole-v1}, where the agent has to balance a pole on a cart that can move to the left and right. The environment starts with the pole close to an upright position and returns a constant reward of $+1$. The episode terminates when the Pole has an angle of $10$ or more degrees to a vertical line or if the cart moves too far to the left or right.

For the dense and shaped reward setting, we use a modified version of \texttt{LunarLander-v2}. In this environment, the agent controls a lunar lander by controlling three engines. The goal is to safely land. The reward contains many components that lead towards the right solution. Among others, the reward includes information about the distance to the ground, the tilting of the lander, and engine usage. However, the original version of \texttt{LunarLander-v0} has two problems. 
\begin{itemize}
    \item The terrain around the landing zone is not observable. In each episode the terrain around the landing zone is randomized, thus the agent can not know when exactly it will crash/land when it's outside the landing zone. This partial observability is problematic for the current version of \method ~as explained in Section \ref{sec:conclusion}.
    \item An episode terminates when the lunar lander object becomes inactive, which is the case when it has not moved for three consecutive frames. However, the observation does not contain information about past motion, therefore the termination condition is also not observable.
\end{itemize}
We slightly modify the problem to obtain \texttt{PredictableLunarLander-v0}, which has a flat moon surface instead of a randomly generated one and where the termination condition is fully dependent on the observation. An episode terminates as a successful landing when both landing feet touch the ground and the velocity is $0$ regardless of past movement.

\subsection{Comparison to \texorpdfstring{$\varepsilon$}{epsilon}-greedy}
\begin{figure}[t]
    \begin{center}
        \includegraphics{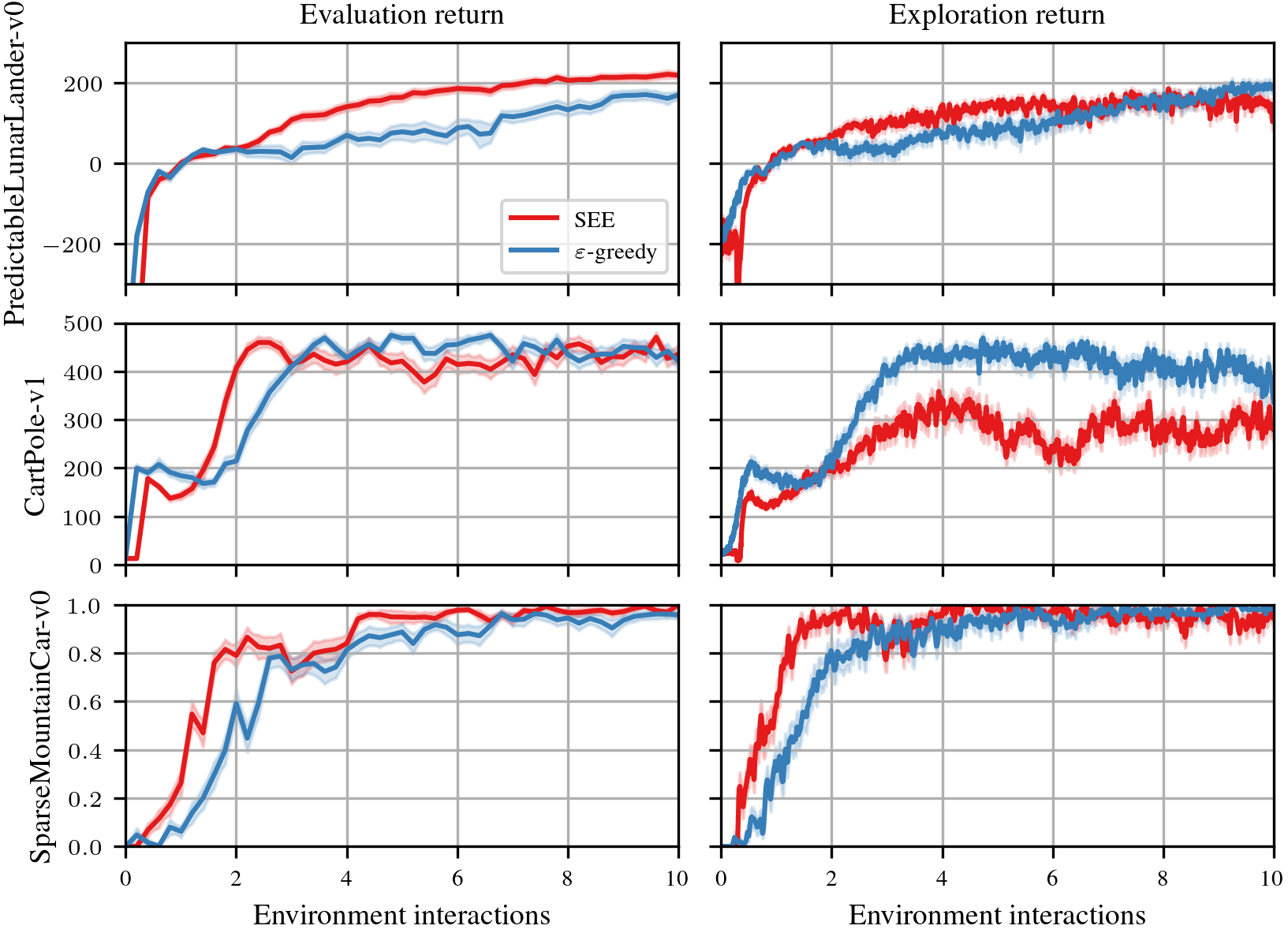}
    \end{center}
    \caption{Comparison between $\varepsilon$-greedy and \method, the x-axis in $10,000$ steps. The line shows the mean performance across $50$ runs, the shaded area shows the standard error of the mean.}
    \label{fig:comparison}
\end{figure}

The considered baseline is identical to the exploitation objective of \method, combining dueling double DQN, but uses the $\varepsilon$-greedy strategy for exploration. Both the baseline and \method~have been separately optimized (see Appendix~\ref{subsec:optimisation}).
The result of this comparison is shown in Figure~\ref{fig:comparison}, showing both the exploration and evaluation returns during training. The evaluation returns are measured by rolling out the policy for $10$ episodes without using any exploration at each $2000$ timesteps during the training, and taking the average across these $10$ runs. The exploration returns are obtained with rollouts generated during training. Data points are generated each time an episode finishes.
\method~outperforms the $\varepsilon$-greedy baseline during training in all environments, with the most pronounced difference being in \texttt{PredictableLunarLander-v0}, the environment with the most shaped reward. Both algorithms handle \texttt{SparseMountainCar-v0} surprisingly well. While the number of updates per environment interaction is fixed, the scheduling when these updates take place is optimized. We found that both algorithms solve \texttt{SparseMountainCar-v0} only when the updates do not take place at each timestep.

\subsection{Ablation study}
\begin{figure}[t]
    \begin{center}
        \includegraphics{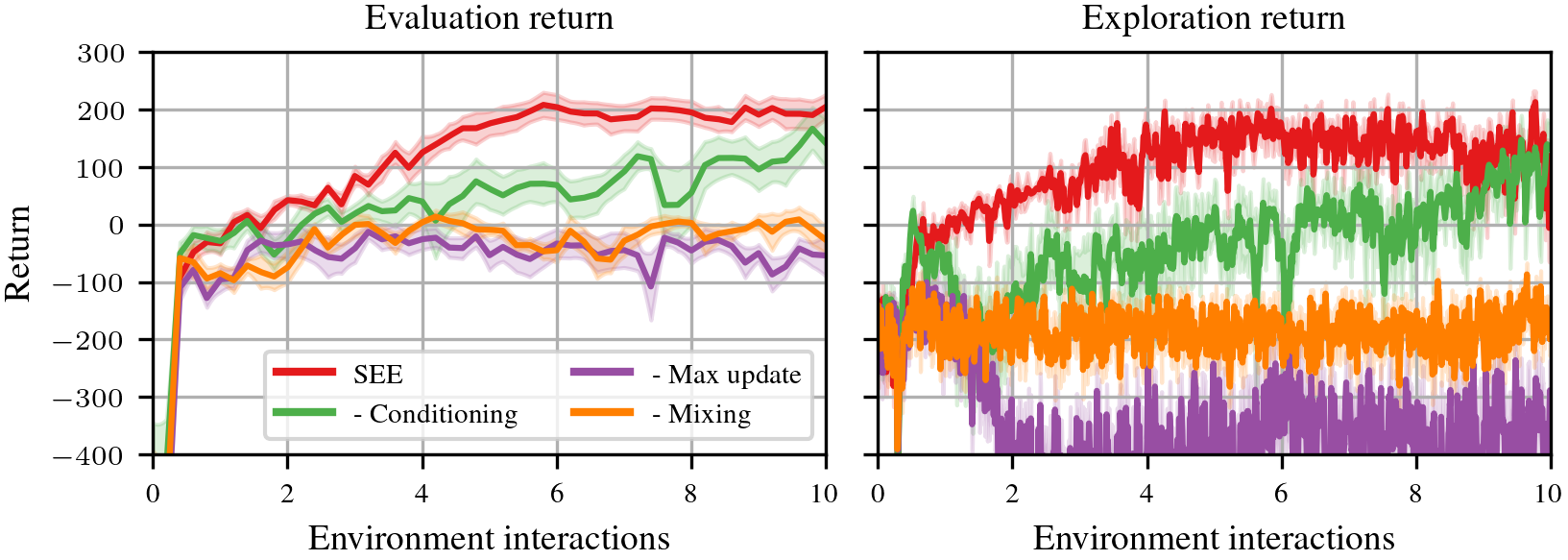}
    \end{center}
    \caption{Comparison of \method~and modified versions where one of the components is replaced or left out. The x-axis is in steps of $10,000$. For this ablation, only the \texttt{PredictableLunarLander-v0} environment has been used. The line shows the mean performance across $10$ runs, the shaded area shows the standard error of the mean.}
    \label{fig:ablation}
\end{figure}

\method~includes three major modifications as described in Section~\ref{sec:method}. To show that each of these components is necessary to achieve the shown result, we conduct an ablation study in the environment with the most pronounced advantage over $\varepsilon$-greedy, i.e., \texttt{PredictableLunarLander-v0} (Figure~\ref{fig:ablation}).
The plot for \method~in Figure~\ref{fig:ablation} is the same as Figure~\ref{fig:comparison}, but only contains the first $10$ of the $50$ runs, to maintain a consistent number of runs across the plot. Every other line excludes one of the three components of \method. For all the runs, we use the same hyperparameters as in Figure~\ref{fig:comparison}.

\textbf{"Conditioning"} removes the Fingerprinting that informs the exploration network about the current state of the exploitation network. This makes the objective non-stationary from the perspective of the exploration objective as there is no information about the TD-error changing for the same state-action pairs. In this version, the TD-error is always calculated on the current parameters of the exploitation network during the exploration update. There is no buffer that holds past exploitation parameters.

\textbf{"Max update"} uses the regular bellman update instead of the maximum update of Equation \eqref{eq:max_update} for the update of the exploration network. We still use the double $Q$-learning variant to calculate the target. We want to note here that the regular Bellman updates might require different hyperparameters to work properly as the scale of the target changes drastically when using this update.

\textbf{"Mixing"} removes the mixing of both objectives during the rollout. Instead, both exploration and exploitation networks are used during the rollout in an alternating fashion. The active network is switched after an episode ends.

This ablation study evinces that each component of \method~is necessary to achieve the best results.

\section{Conclusion and future outlook}
\label{sec:conclusion}
In this paper, we have introduced a new approach for exploration in reinforcement learning~(RL) that frames exploitation and exploration as separate optimization problems. While the exploitation objective is the regular one of RL, we have formulated an exploration objective that maximizes Bellman error to drive exploration, instead of adopting exploration strategies like random sampling, such as $\varepsilon$-greedy. Our approach comes with several components that are necessary to make our method work. We have leveraged a fingerprinting technique to mitigate the non-stationarity induced by the Bellman error maximization and adopted a maximum reward formulation instead of the regular average reward one. We have shown experimentally that our method can outperform $\varepsilon$-greedy on well-known RL problems.

\paragraph{Limitations} Our exploration method relies on the property that the absolute TD-error indicates that there is information to learn in a transition. If the exploitation agent has seen this transition often enough we expect the error to go to zero such that the exploration focuses on other regions. However, this is not the case if the TD-error is caused by an unlearnable transition of the environment for example by stochasticity or partial observability. If the value of the transition cannot be learned, the TD-error will stay constant and the exploration agent would therefore indefinitely seek this transition. This is similar to the "noisy TV" problem~\cite{burda_exploration_2018} but concerning rewards instead of states. Currently, this method can not be successfully applied to most stochastic environments. In the next steps, we will search for solutions to mitigate this issue as well as assessing the validity of \method~in more complex environments and against more sophisticated exploration strategies.

\newpage

\section*{Acknowledgments}
This work was funded by the German Federal Ministry of Education and Research (BMBF) (Project: 01IS22078). The authors gratefully acknowledge the scientific support and HPC resources provided by the Erlangen National High Performance Computing Center (NHR@FAU) of the Friedrich-Alexander-Universität Erlangen-Nürnberg (FAU) under the NHR project b187cb. NHR funding is provided by federal and Bavarian state authorities. NHR@FAU hardware is partially funded by the German Research Foundation (DFG) – 440719683.

\bibliographystyle{unsrtnat} 
\bibliography{main}

\newpage

\appendix

\section{Optimisation procedure}
\label{subsec:optimisation}
For the comparison of Figure~\ref{fig:comparison}, we optimized both DQN with $\varepsilon$-greedy and \method~independently for a fair comparison. In the following, we describe the optimization procedure and report the hyperparameters used for the experiments.
For the optimisation, we used Optuna~\cite{akiba_optuna_2019}. The objective is to maximize the average evaluation return across the three selected environments described in Section~\ref{subsec:environments}. A single trial consists of $9$ runs, $3$ runs per environment. As the environments have different return ranges we need to weigh them such that every environment has roughly the same impact on the results. We aim to normalize the range of returns to the interval $[0, 100]$. For \texttt{SparseMountainCar-v0} and \texttt{CartPole-v1} this is straightforward. Both have a clear range of returns namely $[0,1]$ and $[0, 500]$ respectively. Therefore, we weight them with $100$ and $0.2$ accordingly. \texttt{PredictableLunarLander-v0} does not have a bounded range of returns in either direction. However, a return of $0$ is usually very quickly obtained by learning to hover after a few episodes. According to the Gymnasium documentation~\cite{towers_gymnasium_2023}, the environment is solved at a return of $200$. It is possible to reach a higher return but that is quite hard. Therefore, we opted to treat \texttt{PredictableLunarLander-v0} as if it had a return range of $[0, 200]$ and therefore weight it by $0.5$.\\
To simplify the optimization we disentangled $\tau$, used for the update of the target network, from the update frequency. We optimize "$\tau_\text{per time step}$" and then calculate the $\tau$ used in the configuration together with the update frequency by 
\begin{equation}
    \tau = 1 - (1 - \tau_{\text{per time step}} )^{\text{update frequency}}.
\end{equation}
Since we have fixed the number of gradient steps per environment interaction to $1$, the update frequency equals the number of gradient steps per update. The optimization settings for the optimization of DQN with $\varepsilon$-greedy are reported in Table~\ref{tab:search_dqn}. With this setting, $100$ trials have been executed to find the hyperparameters.
\begin{table}
    \centering
    \begin{tabular}{l|r|r|r|r}
         \textbf{Hyperparameter} & \textbf{Type} & \textbf{Min Value} & \textbf{Max Value} & \textbf{Logarithmic} \\ \hline
         learning rate & float & $0.0001$ & $0.01$ & True \\
         replay buffer size & int & $10000$ & $100000$ & False \\
         warm-up steps & int & $0$ & $10000$ & False \\
         epsilon end & float & $0.0$ & $0.1$ & False \\
         epsilon decay steps & int & $0$ & $50000$ & False \\
         update frequency & int & $1$ & $64$ & False \\
         tau per timestep & float & $0.001$ & $0.01$ & True \\ \hline
    \end{tabular}
    \caption{Optuna hyperparameter search settings for $\varepsilon$-greedy DQN.}
    \label{tab:search_dqn}
\end{table}

\begin{table}
    \centering
    \begin{tabular}{l|r|r|r|r}
         \textbf{Hyperparameter} & \textbf{Type} & \textbf{Min Value} & \textbf{Max Value} & \textbf{Logarithmic} \\ \hline
         exploration discount & float & $0.9$ & $0.99$ & True \\
         mixture & float & $0.0$ & $1.0$ & False \\
         exploration transition batch size & categorical & \multicolumn{3}{c}{$[128, 64, 32, 16, 8, 4, 2, 1]$} \\
         value function replay buffer size & int & $1$ & $128$ & False \\
         fingerprinting probe number of probe states & int & $1$ & $16$ & False \\
         exploitation learning rate & float & $0.0001$ & $0.01$ & True \\
         exploration learning rate & float & $0.0001$ & $0.01$ & True \\
         transition replay buffer & int & $10000$ & $100000$ & False \\
         update frequency & int & $0$ & $64$ & False \\
         exploitation tau per timestep & float & $0.001$ & $0.01$ & True \\
         exploration tau per timestep & float & $0.001$ & $0.01$ & True \\ \hline
    \end{tabular}
    \caption{Optuna hyperparameter search settings for \method.}
    \label{tab:search_seex}
\end{table}

For \method, we use two additional batch sizes to the regular transition batch used for DQN. The two additional batch sizes are the number of transitions and the number of exploitation value function parameters sets $\theta$ from the respective replay buffers. The absolute TD-error is calculated for each transition in combination with each value function set of parameters. Therefore, we have an effective batch size of "exploration transition batch size" $\times$ "value function batch size". To stick to the same effective batch size, as we choose for $\varepsilon$-greedy and the exploitation objective, we optimize the "value function batch size" as a categorical parameter and then choose the exploration transition batch size as
\begin{equation}
    \text{exploration transition batch size} = \frac{\text{batch size}}{\text{value function batch size}}.
\end{equation}
Table \ref{tab:search_seex} shows the search settings used to optimise the hyperparameters of \method. Since there are $11$ hyperparameters compared to only $7$ for $\varepsilon$-greedy, we used $200$ trials to find the optimised hyperparameters.

\section{Hyperparameters selection}

\begin{table}
    \centering
    \begin{tabular}{l|r}
        \textbf{Hyperparameter} & \textbf{Value} \\ \hline
        hidden layer sizes & [$256$, $256$] \\
        discount factor & $0.99$ \\
        batch size & $128$ \\
        gradient clip value & $10$ \\
        gradient steps per environment interaction & $1$ \\ \hline
    \end{tabular}
    \caption{Fixed hyperparameters across methods.}
    \label{tab:fixed_hyper}
\end{table}
\begin{table}
    \centering
    \begin{tabular}{l|r}
         \textbf{Hyperparameter} & \textbf{Value} \\ \hline
         epsilon end & $0.0929$ \\
         epsilon decay steps & $5144$ \\
         learning rate & $0.0004$\\
         replay buffer size & $85317$ \\
         warm-up steps & $194$ \\
         update frequency & $63$ \\
         tau per update & $0.3421$ \\ \hline
    \end{tabular}
    \caption{Optimised hyperparameters for $\varepsilon$-greedy DQN, values are rounded to $4$ decimal points.}
    \label{tab:dqn_hyper}
\end{table}
\begin{table}
    \centering
    \begin{tabular}{l|r}
        \textbf{Hyperparameter} & \textbf{Value} \\ \hline
        exploration transition batch size & $4$ \\
        value function batch size & $32$ \\
        exploration discount & $0.9724$ \\
        exploitation tau & $0.1700$ \\
        exploration tau & $0.1622$ \\
        exploitation learning rate & $0.0007$ \\
        exploration learning rate & $0.00851$ \\
        transition replay buffer size & $16517$ \\
        value function replay buffer size & $2$ \\
        fingerprinting number of probe states & $12$ \\
        warm-up steps & $2829$ \\
        mixture factor & $0.3525$\\
        update frequency & $21$ \\ \hline
    \end{tabular}
    \caption{Optimised hyperparameters for \method, values are rounded to $4$ decimal points.}
    \label{tab:seex_hyper}
\end{table}

Table \ref{tab:dqn_hyper} and Table \ref{tab:seex_hyper} show the optimized hyperparameters for DQN with $\varepsilon$-greedy and \method\, respectively. These are the hyperparameters that have been used for the experiments. Note that there is a nonsensical combination of hyperparameters for \method. The "value function batch size" is $34$, while the value function replay buffer size is only $2$. This could either mean that more search is necessary to find good hyperparameters, that the method requires more gradient steps, or that we are misunderstanding a property of the method and its hyperparameters.
Also, note that the hyperparameters of DQN are all within expected ranges. Calculating the tau per timestep here gives $0.0066$ which is close to the often-used default value of $0.005$.

\end{document}